\pdfoutput=1

\documentclass[sigconf, nonacm]{acmart}

\usepackage{latexsym}
\usepackage{subfigure}
\usepackage{multirow}
\usepackage{listings}
\AtBeginDocument{%
  \providecommand\BibTeX{{%
    \normalfont B\kern-0.5em{\scshape i\kern-0.25em b}\kern-0.8em\TeX}}}

\setcopyright{none}
\renewcommand\footnotetextcopyrightpermission{}
\begin{document}

\title{Thinking with Knowledge Graphs: Enhancing LLM Reasoning Through Structured Data}

\author{Xue Wu}
\affiliation{%
  \institution{Yahoo Research}
  \city{Mountain View}
  \state{California}
  \country{USA}
}
\email{xuewu@yahooinc.com}

\author{Kostas Tsioutsiouliklis}
\affiliation{%
  \institution{Facts.ai}
  \city{Saratoga}
  \state{California}
  \country{USA}
}
\email{kostas@facts.ai}

\begin{abstract}
Large Language Models (LLMs) have demonstrated remarkable capabilities in natural language understanding and generation. However, they often struggle with complex reasoning tasks and are prone to hallucination. Recent research has shown promising results in leveraging knowledge graphs (KGs) to enhance LLM performance. KGs provide a structured representation of entities and their relationships, offering a rich source of information that can enhance the reasoning capabilities of LLMs. For this work, we have developed different techniques that tightly integrate KG structures and semantics into LLM representations. Our results show that we are able to significantly improve the performance of LLMs in complex reasoning scenarios, and ground the reasoning process with KGs. We are the first to represent KGs with programming language and fine-tune pretrained LLMs with KGs. This integration facilitates more accurate and interpretable reasoning processes, paving the way for more advanced reasoning capabilities of LLMs.

\end{abstract}


\maketitle

\section{Introduction}

Large Language Models (LLMs) have achieved state of art performance in many Natural Language Processing (NLP) tasks (\cite{brown2020languagemodelsfewshotlearners}, \cite{chowdhery2022palmscalinglanguagemodeling}). They have been successfully applied in a wide range of applications, from question answering to summarization and machine translation. However, due to limitations in the training data and training process, they suffer from hallucination \cite{Ji_2023}, where generated text is nonfactual, nonsensical, or incoherent. This issue becomes particularly prevalent when dealing with tasks that require intricate or complex reasoning. To address hallucination, researchers have used methods such as prompting (\cite{wei2022chainofthought}, \cite{zhang2024chainpreferenceoptimizationimproving}, \cite{wang2023selfconsistency}, \cite{fu2023complexitybased}, \cite{pressMeasuringNarrowingCompositionality2023}, \cite{yao2023react}, \cite{yao2023tree}), retrieval augmented generation (RAG \cite{lazaridou2022internetaugmentedlanguagemodelsfewshot}), and fine-tuning. These approaches often leverage external sources of information, which can come from the internet, third-party applications, databases, or knowledge graphs. 

Knowledge Graphs (KGs) are structured representations of real world entities and the relationships among them, offering a rich source of factual information (\cite{Lehmann2014DBpedia} \cite{Bollacker2008FreebaseAC}). By grounding the reasoning processes of LLMs with KGs, we can enhance the factual accuracy of the generated text and reduce hallucinations. Researchers have explored various approaches to integrate KGs with LLMs, each with its own advantages and limitations. One approach involves using Graph Neural Networks (GNNs) to encode KGs into embeddings that capture the structural and semantic information of the graphs (\cite{Choudhary2022}, \cite{yasunaga2022deepbidirectionallanguageknowledgegraph}). These embeddings then serve as soft prompts to LLMs, guiding the generation process with knowledge from the KGs. However, this method requires very careful design and tuning, as it needs to align the representations learned by GNNs with the token-based processing of LLMs. Moreover, since such integration is specific per graph, it may require significant re-engineering for different tasks or graph types.

Another method uses semantic parsing to convert natural language queries into structured query languages like SPARQL (\cite{xu-etal-2023-fine}, \cite{brei2024leveragingsmalllanguagemodels}, \cite{rangel2024sparqlgenerationanalysisfinetuning}, \cite{bustamante2024sparqlgenerationentitypretrained}). In this approach, the LLM generates a SPARQL query based on the input prompt, which is then executed against the KG to retrieve relevant information. This method effectively uses the KG as an external knowledge base, and treats the KG and the LLM as separate components. As a result, the reasoning process is not fully integrated into the LLM, potentially limiting the model's ability to perform complex reasoning during text generation.

Alternatively, researchers encode entities and relationships of KGs as natural language text (\cite{luo2024reasoninggraphsfaithfulinterpretable}, \cite{dernbach2024glamfinetuninglargelanguage}). They either incorporate them into the LLM's input context or fine-tune the LLM with these text representations. This approach leverages the LLM's natural language understanding ability to reason over the text-encoded knowledge. However, representing structured data as unstructured text poses challenges. Capturing the nuances of entities and relationships in natural language requires careful design to avoid ambiguities and ensure that the structural information is preserved, which can be non-trivial for complex graphs.

While there are different ways to leverage KGs to enhance LLMs' performance, there are two aspects to using KGs for improving LLMs. One is grounding the LLMs with trustworthy information, and the other is providing them with examples of relations from which they can generalize. It is this second part that motivated our work. In this work, we propose an approach that represents knowledge graphs with programming language code. Programming languages are inherently structured and are designed to represent complex data and relationships efficiently. This allows for an accurate encoding of entity relationships that preserves the internal structure of the graph. More importantly, programming code is part of the pre-training data for many LLMs, meaning that the models are already equipped to parse and understand programming syntax and semantics. This reduces the need for additional specialized training to interpret the KG representations. By leveraging the structured representation of KGs in programming languages, LLMs can perform more sophisticated reasoning over the data. We investigate different methods of integrating entity relationships into LLMs. Our experimental results demonstrate that programming language (Python) representations of KGs outperform traditional natural language representations and structured JSON representations in complex reasoning tasks, leading to more accurate and reliable outputs.

The main contributions of this paper are:
\begin{itemize}
    \item We introduce a novel representation of knowledge graphs with programming language. It facilitates the seamless integration of structured knowledge into the language modeling process.
    \item By tightly integrating knowledge graphs into LLMs, our approach improves the reasoning accuracy of LLMs on complex tasks and effectively grounds the reasoning process, reducing the chance for hallucinations.
\end{itemize}

\section{Related Work}

There have been several attempts to apply LLMs to graph reasoning tasks. Wang et al. \cite{wang2024languagemodelssolvegraph}, Guo et al. \cite{guo2023gpt4graphlargelanguagemodels}, and Ye et al. \cite{ye2024languagegraphneeds} employ the Graph2Text strategy of converting graph data into textual descriptions. However, these textual descriptions can result in very large contexts, and algorithms such as shortest path computations and bipartite graph matching require calculations across the entire context, making the task highly challenging. Chai et al. \cite{chai2023graphllmboostinggraphreasoning} have introduced GraphLLM, which combines three steps, namely node understanding, graph structure understanding, and graph-enhanced prefix tuning. Zhu et al. \cite{zhu2024investigatinginstructiontuninglarge}, and Wang et al. \cite{wang2024instructgraphboostinglargelanguage} proposed different methods for instruction fine-tuning LLMs to impove the performance of common graph tasks.

While the above works address general graph problems, other research has focused specifically on combining KGs with LLMs.
One such approach is to use the LLM as an encoder to transform text-based entity nodes and relations, and then fuse the LLM and GNN-derived representations. Applications of this approach range from product recommendation (Choudhary et al. \cite{Choudhary2022}) to biomedical question-answering (Yasunaga et al. \cite{yasunaga2022deepbidirectionallanguageknowledgegraph}).
Luo \& Pan \cite{luo2024reasoninggraphsfaithfulinterpretable} have proposed a reasoning on Graph (RoG) method that comprises two modules: a planning module and a retrieval-reasoning module. The planning module mines the KG and generates faithful relation paths for answering the question. The retrieval-reasoning module combines a breadth-first search and a probabilistic optimization over all paths. Dernbach et al. \cite{dernbach2024glamfinetuninglargelanguage} developed a neighborhood partitioning and encoding scheme to accommodate real-world graph properties. Their encoding scheme transforms graph relations into alternate text representations, which in turn are used to fine-tune the LLM. 

Edge et al. \cite{edge2024localglobalgraphrag} have built an end-to-end system, called GraphRAG, that starts with a set of source documents and iteratively applies an LLM to extract the entities in each document. Next, entities are connected via the relationships extracted, and a knowledge graph is created. The knowledge graph is split into communities, and each community is summarized. These summaries are subsequently used by the LLM via RAG to help answer questions submitted to the system.

Nie et al. \cite{nie2024codestyleincontextlearningknowledgebased} provide a code-style in-context learning (ICL) method for knowledge base question answering (KBQA). They design seven meta-functions written in Python that cover the atomic operations used for querying databases. By using few-shot learning, they improve LLMs' ability to query knowledge bases effectively.

There is also ongoing research work (\cite{ma2023trainingstagedoescode}, \cite{zhang2024unveilingimpactcodingdata}, \cite{aryabumi2024codecodeexploringimpact}) that studies the impact of mixing programming code in pre-training or instruction fine-tuning datasets on the performance of LLMs. Even though the programming code used by the research is generic, the results show promising improvements on various tasks, including reasoning tasks. There is another line of work (\cite{kim2023cotcollectionimprovingzeroshot}, \cite{zhang2024chainpreferenceoptimizationimproving}) that studies how to improve LLM Chain of Thought (CoT) reasoning ability through fine-tuning. However, creating good datasets for fine-tuning CoT is labor intensive, and the reasoning steps are described in natural language text, which can introduce ambiguity.

Yang et al. \cite{yang2024largelanguagemodelslatently} studied how LLMs do multi-hop reasoning and found that LLMs perform latent multi-hop reasoning in certain relation types, but often struggle with later hops.
Biran et al. \cite{biran2024hoppinglateexploringlimitations} observed that later hops are resolved in the model's deeper layers, and thus the LLM may no longer encode the necessary knowledge for answering the question. They propose a back-patching approach, essentially feeding the hidden representations from later layers back into earlier ones.

In this work, we continue the prior research work on multi-hop queries, showing that we can significantly improve the performance of LLMs by integrating KG structures and semantics into LLM representations. Similarly to \cite {nie2024codestyleincontextlearningknowledgebased} we also experiment with code-style representations. However, in our case, we represent the entity-relations -not the atomic operations- in Python, and our goal is to improve the LLMs' ability to answer questions directly, not its ability to query the knowledge graph.

The rest of the paper is structured as follows. In Section~\ref{methodology} we describe the methodologies we followed to prompt or fine-tune the LLM. In Section~\ref{experiments} we describe the experimental design and results. Finally, we conclude the paper in Section~\ref{conclusion}.

\section{Methodology} \label{methodology}
Our work focuses on studying the entity relationships representation of KG for grounded LLM reasoning. 
\subsection{Knowledge Graph Definition}
Let $G=\{E, R, T\}$ denote a knowledge graph, where $E$ is the set of entities, $R$ is the set of relationships, $T\subseteq E\times R\times E$ is the set of triplets that are edges in the knowledge graph. A triplet $(e_i, r_i, e_{i+1}) \in T$ if there is a directed edge between entity $e_i$ ($e_i \in E$) and entity $e_{i+1}$ ($e_{i+1} \in E$) through relationship $r_i$ ($r_i \in R$). A triplet also corresponds to a complete one hop reasoning process. A two hop compositional reasoning process can be represented as $((e_i, r_i, e_{i+1}), (e_{i+1}, r_{i+1}, e_{i+2}))$, where $(e_i, r_i, e_{i+1}) \in T$ and $(e_{i+1}, r_{i+1}, e_{i+2}) \in T$. Since $e_{i+1}$ exists in both triplets, it is the bridge entity. Similarly, a three hop compositional reasoning process can be represented as $((e_i, r_i, e_{i+1}), (e_{i+1}, r_{i+1}, \allowbreak e_{i+2}), (e_{i+2}, r_{i+2}, e_{i+3}))$, where $(e_i, r_i, e_{i+1}) \in T$, $(e_{i+1}, r_{i+1}, e_{i+2}) \in T$ and $(e_{i+2}, r_{i+2}, e_{i+3}) \in T$.

\subsection{Knowledge Graph Representation for LLM Reasoning}
To improve LLM multihop reasoning with knowledge graphs, we represent knowledge graphs in ways that are more compatible with LLM prompting and fine-tuning. When given a complex reasoning prompt, LLMs can detect entities and relationships, then implicitly infer the key entities and facts by following logical reasoning steps grounded by knowledge graphs. For instance, given the prompt: ``Who is the spouse of the composer of `It Goes Like It Goes'?'', LLMs can follow the reasoning steps: ``The composer of `It Goes Like It Goes' is David Shire'', and ``The spouse of David Shire is Didi Conn", and infer that the correct answer is ``Didi Conn''. Different representations of knowledge graphs affect how effectively LLMs can perform such logical reasoning.

\begin{figure}[htbp]
    \centering
    \begin{minipage}{\columnwidth}
        \centering
        \includegraphics[width=0.9\textwidth]{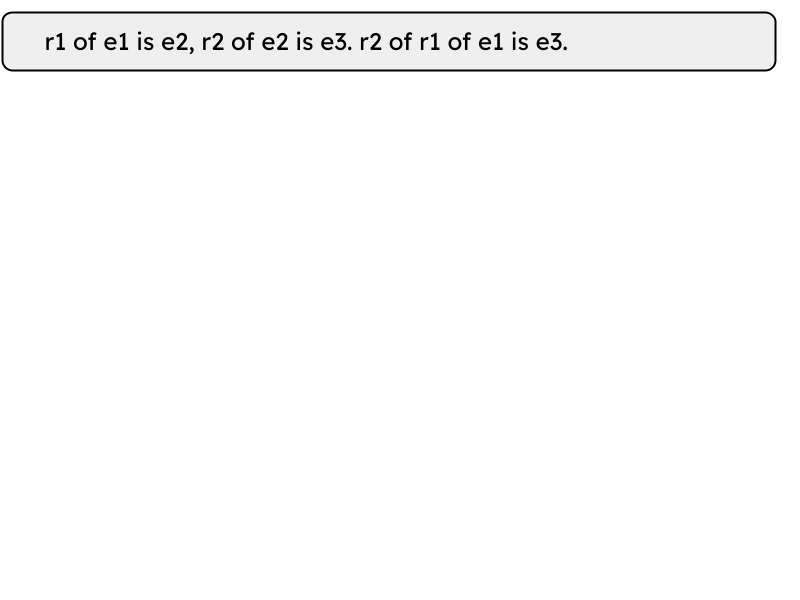}
        \vspace{-4.8cm}
        \caption{Natural Language Representation of KG with Static Relationships}
        \label{fig:natural-language}
        \vspace{0.4cm}
    \end{minipage}    

    \begin{minipage}{\columnwidth}
        \centering
        \includegraphics[width=0.9\textwidth]{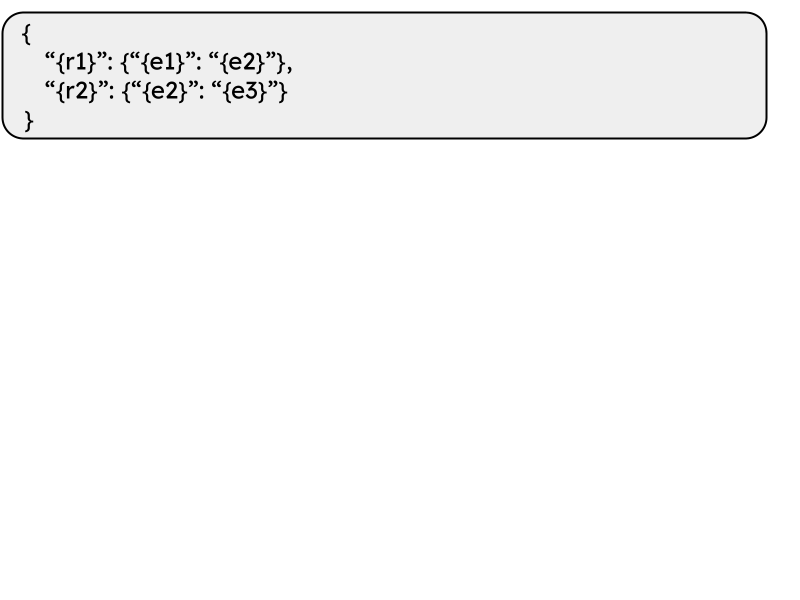}
        \vspace{-4.2cm}
        \caption{JSON Representation of KG with Static Relationships}
        \label{fig:json}
        \vspace{0.4cm}
    \end{minipage} 

    \begin{minipage}{\columnwidth}
        \centering
        \includegraphics[width=0.9\textwidth]{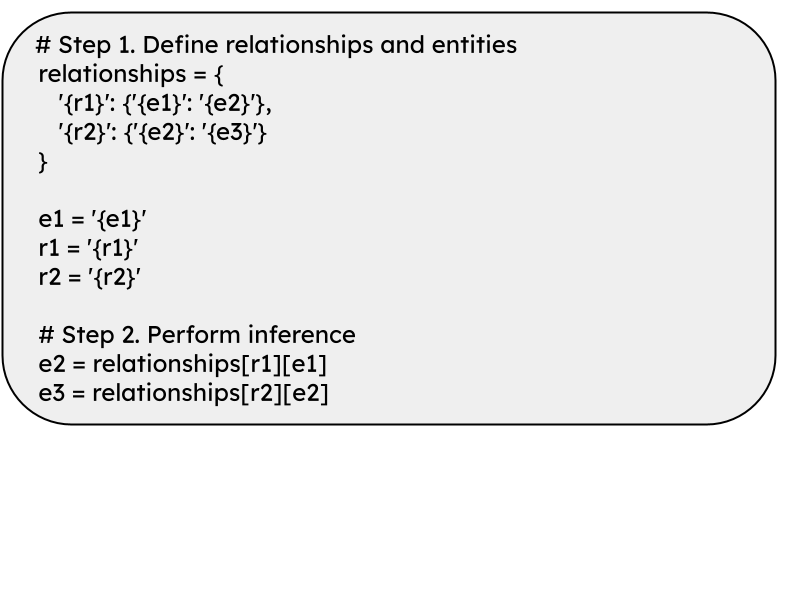}
        \vspace{-1.4cm}
        \caption{Python Representation of KG with Static Relationships}
        \label{fig:python-v1}
        \vspace{0.4cm}
    \end{minipage}    

    \begin{minipage}{\columnwidth}
        \centering
        \includegraphics[width=1.1\textwidth]{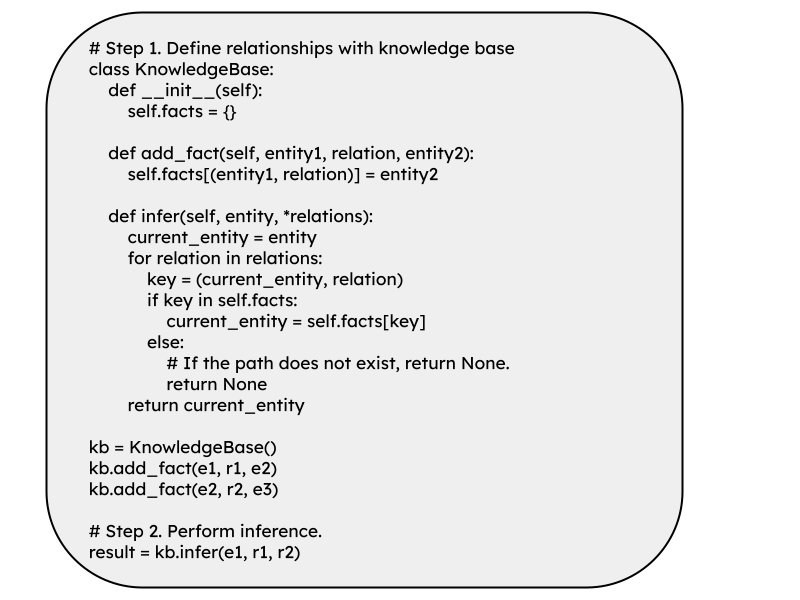}
        \caption{Python Representation of KG with Dynamic Relationships}
        \label{fig:python-v2}
        \vspace{-0.1cm}
    \end{minipage}      
\end{figure}

The most natural way is to use natural language to describe the triplets in knowledge graphs. Figure \ref{fig:natural-language} shows the natural language representation for two-hop reasoning of triplets $((e_1, r_1, e_2), (e_2, r_2,\allowbreak e_3))$, where $e_2$ is the bridge entity. For instance, the triplets ((`It Goes Like It Goes', `composer', `David Shire'), (`David Shire', `spouse', `Didi Conn')) can be represented as (The composer of `It Goes Like It Goes' is David Shire, The spouse of David Shire is Didi Conn). And the two hop reasoning can be represented as ``The spouse of the composer of `It Goes Like It Goes' is Didi Conn''. As LLMs understand natural languages very well, the natural language representation is the most straightforward way for either prompting or fine-tuning LLMs with knowledge graphs. 

JSON (JavaScript Object Notation) is a lightweight data interchange format \cite{bray2014javascript}. It is designed to store data in universal data structures such as dictionary and list that are supported by most programming languages. JSON is a pure data only format and can be used to store structured data from knowledge graphs. Figure \ref{fig:json} shows the JSON representation of knowledge graph triplets $(e_1, r_1, e_2)$ and $(e_2, r_2, e_3)$. The entities $e_1$ and $e_2$ are the keys and the relationship/entity pairs $r_1:e_2$ and $r_2:e_3$ are the values. However, since JSON is designed to store data only, it is difficult to represent multi-hop inference process with the JSON format. Alternatively, a comment or description field (with ``comment'' as the key and natural language description of the multi-hop reasoning as the value) can be added to the JSON representation. But this is not a recommended practice in general.

Programming language code is another major data source for LLM pre-training and fine-tuning. Knowledge graphs can be represented using various data structures supported by major programming languages such as Python. The triples can be represented either as a static dictionary or added dynamically to the predefined data structures as part of the code. Figure \ref{fig:python-v1} shows Python representation of knowledge graph triplets with a static dictionary data structure, and the two-hop inference process that is based on the stored triplets. As shown in figure \ref{fig:python-v1}, relationships and entities are stored in dictionary data structure ``relationships'' with relationships $r_1$ and $r_2$ as the keys and entity pairs as the values. The inference process is just the process of retrieving the values based on the key values of the dictionary. Figure \ref{fig:python-v2} shows Python representation of knowledge graph triplets with predefined Python class ``KnowledgeBase'', and an iterative multi-hop inference function `infer' that supports inference of an arbitrary number of hops. As shown by figure \ref{fig:python-v2}, the main data structure of ``KnowledgeBase'' is a dictionary ``self.facts'', and entities and relationships are added to the dictionary with function ``add\_fact''. The ``infer'' function accepts any number of relationships with parameter ``*relations'' and does the corresponding multi-hop reasoning. We designed the dynamic self-defined data structure based Python representation because it can be easily generalized to support multi-hop reasoning based on subgraphs of KGs. 

\begin{figure*}[!htbp]
\centering
\begin{minipage}{0.31\linewidth}
    \begin{minipage}{\columnwidth}
        \centering
        \includegraphics[width=1.0\linewidth]{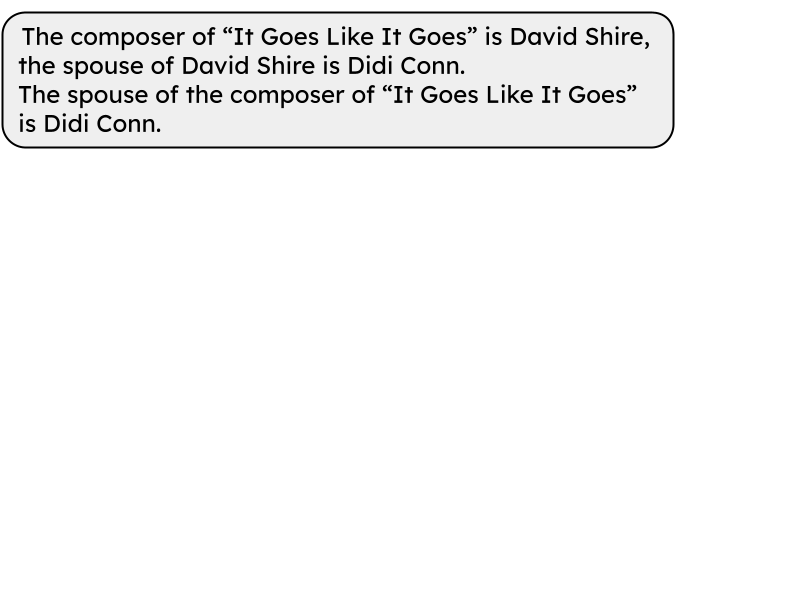}
        \label{fig:nl-example}
        \vspace{-2.5cm}
    \end{minipage}   
    \vspace{-3.5cm}
    \begin{minipage}{\columnwidth}
        \centering
        \includegraphics[width=1.0\linewidth]{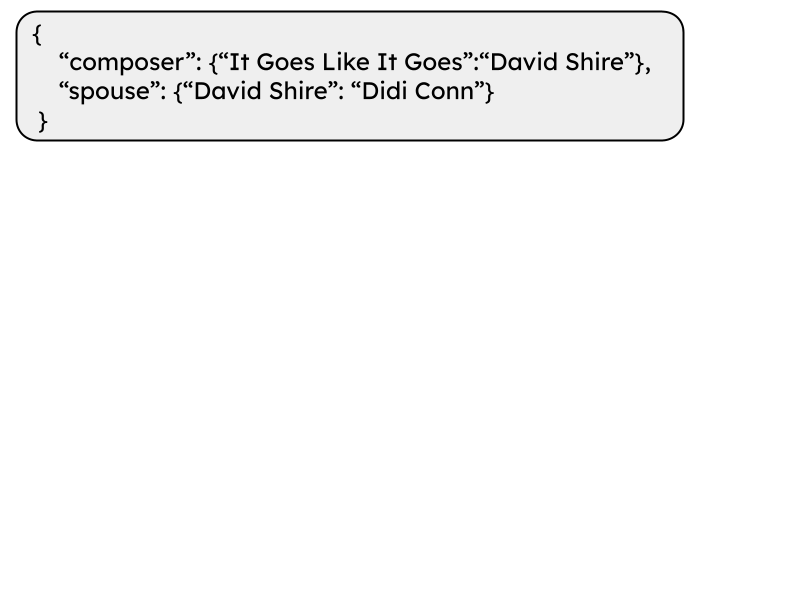}
        \label{fig:json-example}  
    \end{minipage}  
    \vspace{1.5cm}
\end{minipage}
\begin{minipage}{0.31\linewidth}
  \includegraphics[width=1.1\linewidth]{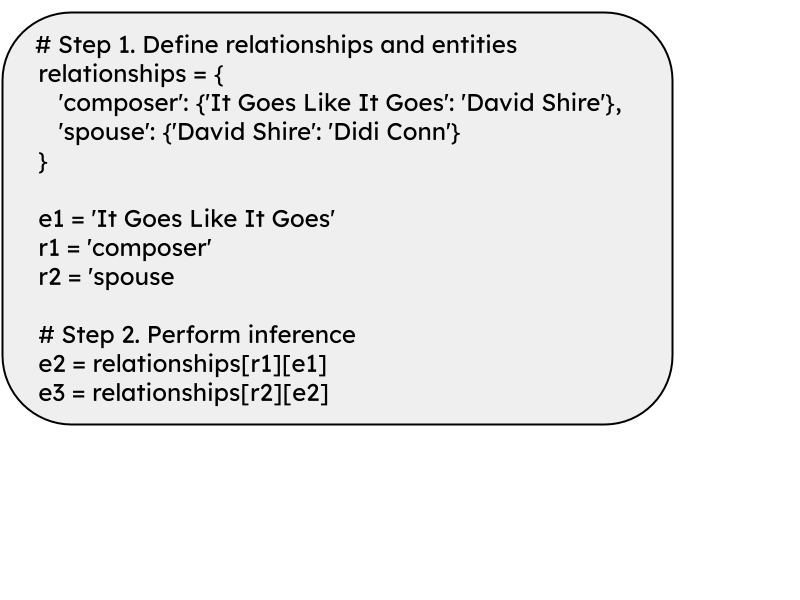}
  \label{fig:python-v1-example}
\end{minipage}
\begin{minipage}{0.31\linewidth}
  \includegraphics[width=1.1\linewidth]{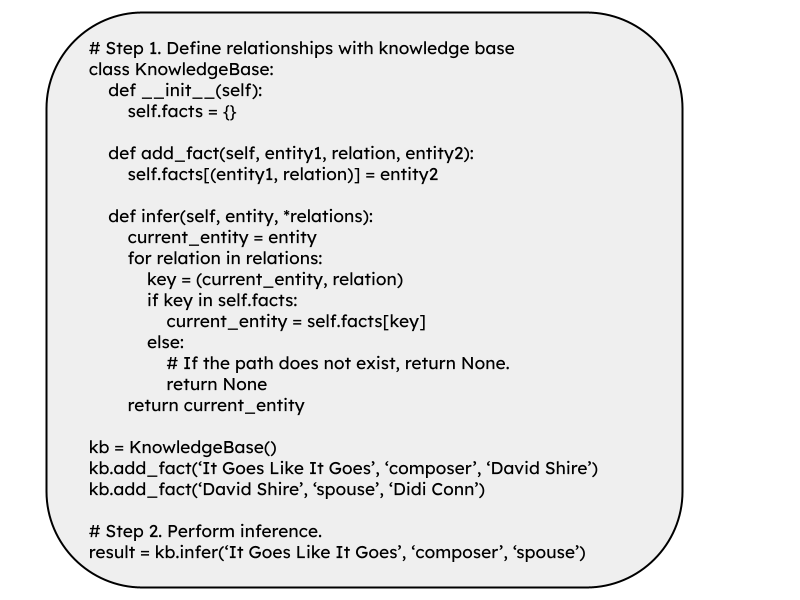}
  \label{fig:python-v2-example}
\end{minipage}
\vspace{-0.3cm}
\caption{Examples of Different Representation of KG}
\label{fig:examples-representation}
\end{figure*}

The example mentioned in the previous paragraphs can be easily represented as dictionaries in Python. Figure \ref{fig:examples-representation} shows four different representations for the same example. Using programming languages, knowledge graphs can be represented in a more controlled and unambiguous way; corner cases can be easily checked. However, the representation for the same triplet can be more verbose. 

\section{Experiments} \label{experiments}
We designed experiments to study how different representations of entity relationships in KGs affect the reasoning performance of LLMs across two different datasets.
\subsection{Datasets}
\subsubsection{Dataset 1} \label{dataset1}
For the first dataset, we use the same dataset used by the ``Hopping Too Late'' paper (\cite{biran2024hoppinglateexploringlimitations}). The dataset includes two-hop relationships extracted from publicly available knowledge base Wikidata (\cite{vrandevcic2014wikidata}). For the experiments, we split the dataset into eight equal-sized partitions based on the bridge entity $e_2$ in a round-robin fashion, so that any unique $e_2$ exists only in one partition. We partition the dataset in this way so that the LLMs only learn the relationships and logical reasoning process rather than memorizing the entities. To avoid overrepresentation of the most popular $e_2$ in the training or testing dataset, we choose partition 2 as our training dataset and partition 4 as the testing dataset. The details of the dataset are listed in Table \ref{tb:dataset1_train_test_split}. For this dataset, there is no overlap of bridge entities between training and testing dataset. The overlap of relationship pairs $(r_1, r_2)$ is about $99\%$.

\begin{table}[t]
\centering
\small
\begin{tabular}{lccc}
\hline
 & & & Train and Test \\
 & Train & Test & Intersection \\
\hline
Number of Hops & 2 & 2 & 2 \\
Dataset Size & 10,262 & 10,255 & 0 \\
Bridge Entities ($e_2$) & 324 & 880 & 0 \\
Relations ($r_1$, $r_2$) & 191 & 207 & 167 \\
No. of row with ($r_1$, $r_2$) & 10,262 & 10,255 & 10,130 \\
\hline
\end{tabular}
\vspace{1em}
\caption{Dataset 1 Train and Test Data Selection}
\label{tb:dataset1_train_test_split}
\vspace{-0.5em}
\end{table}

\subsubsection{Dataset 2} \label{dataset2}
For the second dataset, we use the dataset created in paper (\cite{xanh2020_2wikimultihop}). This dataset also includes two-hop relationships extracted from publicly available knowledge base Wikidata (\cite{vrandevcic2014wikidata}). Although both Dataset 1 and Dataset 2 are derived from the same knowledge base, the extracted entities and relationships for Dataset 2 are different from those in Dataset 1. For the training dataset, we select only compositional relationships from this dataset and limit the number of instances per relationship pair $(r_1, r_2)$ to no more than 500 to avoid over representation of particular relationships in the training data. We use the development dataset with compositional relationships for testing purpose. The details of the dataset are listed in Table \ref{tb:dataset2_train_test_split}. We choose compositional relationships so the type of relationships are consistent with those in Dataset1 and it is easier to compare the reasoning performance across datasets. We didn't further restrict the entities and relationships based on the overlap between training and testing dataset to respect the design of the dataset by paper (\cite{xanh2020_2wikimultihop}).

\begin{table}[t]
\centering
\small
\begin{tabular}{lccc}
\hline
 & & & Train and Test \\
 & Train & Test & Intersection \\
\hline
Number of Hops & 2 & 2 & 2 \\
Dataset Size & 10,733 & 5,236 & 0 \\
Bridge Entities ($e_2$) & 4,170 & 3,650 & 782 \\
Relations ($r_1$, $r_2$) & 80 & 110 & 66 \\
No. of row with ($r_1$, $r_2$) & 10,733 & 5,236 & 1,907\\
\hline
\end{tabular}
\vspace{1em}
\caption{Dataset 2 Train and Test Data Selection}
\label{tb:dataset2_train_test_split}
\vspace{-0.5em}
\end{table}

\subsubsection{Dataset 3} \label{dataset3}
This dataset is an extension of Dataset 1. We extended two-hop relationships $((e_1, r_1, e_2), (e_2, r_2, e_3))$ by adding a third hop $(e_3, r_3, e_4)$, resulting in three-hop relationships $((e_1, r_1, e_2), \allowbreak (e_2, r_2, e_3), (e_3, r_3, e_4))$, while keeping the entities $e_3$ as a subset of the entities $e_3$ in Dataset 1. This dataset was created to test whether models fine-tuned for two-hop reasoning can generalize to improve three-hop reasoning performance as well. The details of the dataset are listed in Table \ref{tb:dataset3_info}. The overlap of bridge entities between training and testing dataset is minimal for this dataset. There is high percentage of relationship overlap between training and testing dataset.

\begin{table}[t]
\centering
\small
\begin{tabular}{lccc}
\hline
 &  & & Intersection with \\
 & Dataset 1 Train & Test & Dataset 1 Train \\
\hline
Number of Hops & 2 & 3 & 0 \\
Dataset Size & 10,262 & 1,007 & 0 \\
Bridge Entities e2 & 324 & 261 & 28 \\
Train $e_2$ vs Test $e_3$ & 324 & 233 & 15 \\
Relations ($r_1$, $r_2$) & 191 & 81 & 70 \\
No. of row with ($r_1$, $r_2$) & 10,262 & 1,007 & 973 \\
Relations ($r_2$, $r_3$) & *191 & 97 & 92 \\
\hline
\end{tabular}
Note *: this number is for ($r_1$, $r_2$) in Dataset 1 Train
\vspace{1em}
\caption{Dataset 3 Basic Statistics}
\label{tb:dataset3_info}
\vspace{-1.5em}
\end{table}

The final training data for fine-tuning the LLMs include both one-hop prompts and responses based on the first hop from the datasets, and two-hop prompts and responses based on two-hop information from the datasets.

\subsection{Large Language Models}
To evaluate the performance of LLM reasoning, we chose the latest released open source models by Meta: Llama-3.1-8B-Instruct and Llama-3.1-70B-Instruct (\cite{dubey2024llama3herdmodels}). The Llama-3.1 model family demonstrates stronger reasoning abilities compared to other open-source LLMs. For our experiments, we fine-tuned the smaller model (Llama-3.1-8B-Instruct) with different entity relationship representations of multi-hop reasoning. During fine-tuning, we optimize the language model objective, which predicts the probability distribution of each token based on its previous tokens. We adopt LoRA (\cite{hu2022lora}) to perform parameter-efficient fine-tuning. During inference, we used greedy decoding for reproducibility.

\subsection{Experiment Environment}
We use the large language model checkpoints from Hugging Face Transformers\footnote{\url{https://huggingface.co/docs/transformers/index}} in all the experiments. Experiments were conducted using one Nvidia A100 processor with 40GB of RAM per GPU. When fine-tuning the LLMs, we ran the training process for one epoch across all the experiments. We didn't fine-tune the larger model (Llama-3.1-70B-Instruct) because of the computation resource constraint. 

\subsection{Experiment Design}
To test how well each representation of KGs helps with LLM reasoning, we designed the following three types of experiments. 
\subsubsection{LLM Reasoning without Context}
To test the multi-hop reasoning ability of LLMs, we compared model performance under three conditions:
\begin{enumerate}
    \item Zero shot prompting of the LLMs to predict $e_3$. Given $(e_1, r_1, r_2)$, we prompt the LLMs to predict $e_3$. This serves as our baseline experiment. 
    \item One shot prompting of the LLMs with reasoning example. We have conducted three experiments, one for each representation format of KGs. 
    \item Zero shot prompting of fine-tuned LLMs. We have conducted three experiments, one for each representation format of KGs. 
\end{enumerate}

\subsubsection{LLM Reasoning with Context}
Since Retrieval Augmented Generation (RAG) is one of the most popular applications of LLMs, we designed experiments to compare the performance of LLMs when given input context. This set of experiments are designed to test how well LLMs perform for potential RAG applications.

\subsubsection{Model Generalization to Longer Reasoning Path}
We fine-tuned the LLMs using only one-hop and two-hop triplets. For this set of experiments, we tested how well the fine-tuned models can generalize to reasoning over longer paths.

\subsection{Prompt Design}
The design of the prompts for all the experiments is presented in Table \ref{tb:prompt-design}. Examples are shown for two-hop reasoning prompts. Because of the paper length limitation, the natural language explanation, JSON data structure and Python code snippets for one-shot prompting are shown in Appendix \ref{prompt-format}. Three-hop reasoning prompts follow the same logic and use the same prompt format. 
\begin{table*}[t]
\centering
\small
\begin{tabular}{lll}
\hline
Dataset & Dataset1 & Dataset2 \\
\hline
Multihop Question & $r_2$ of $r_1$ of $e_1$ is & What is $r_2$ of $r_1$ of $e_1$ ? \\
\hline
Zero Shot Prompt & Given the incomplete statement: $r_2$ of $r_1$ of $e_1$ is \_ , & Given the question: What is $r_2$ of $r_1$ of $e_1$ ?  \\
& provide answer and generate explanation & generate explanation and provide answer to the question \\
& for completing the statement & \\
 \hline
One Shot Prompt & Given the incomplete statement: $r_2$ of $r_1$ of $e_1$ is \_  , & Given the question:  What is $r_2$ of $r_1$ of $e_1$ ? \\
for text representation & \{``Answer”:``$e_3$'', ``Explanation”:``$r_1$ of $e_1$ is $e_2$. $r_2$ of $e_2$ is $e_3$. & \{``Answer”:``$e_3$'', ``Explanation”'':``$r_1$ of $e_1$ is $e_2$. $r_2$ of $e_2$ is $e_3$. \\
  & $r_2$ of $r_1$ of $e_1$ is $e_3$''\} &  $r_2$ of $r_1$ of $e_1$ is $e_3$''\} \\
 & Given the incomplete statement: \{statement\} \_  , & Given the question: \{question\} ? \\
 & provide answer and generate explanation & generate explanation and provide answer to the question \\ 
 & for completing the statement & generate explanation and provide answer to the question \\
Or for JSON & Given the incomplete statement: $r_2$ of $r_1$ of $e_1$ is \_  , & Given the question:  What is $r_2$ of $r_1$ of $e_1$ ? \\
representation & \{``Answer”:``$e_3$”, ``JSON Structure'':``\{JSON data\}'' & \{``Answer”:``$e_3$'', ``JSON structure'':``\{JSON data\}'' \\
 & Given the incomplete statement: \{statement\} \_  , & Given the question: \{question\} ? \\
 & provide answer and generate JSON structure & generate JSON structure and provide answer to the question \\ 
 & for completing the statement &  \\  
Or for Python & Given the incomplete statement: $r_2$ of $r_1$ of $e_1$ is \_  , & Given the question:  What is $r_2$ of $r_1$ of $e_1$ ? \\
representation & \{``Answer”:``$e_3$”, ``Python code snippet”:``\{Python code\}" & \{``Answer”:``$e_3$”, ``Python code snippet”:``\{Python code\}" \\
 & Given the incomplete statement: \{statement\} \_  , & Given the question: \{question\} ? \\
 & provide answer and generate python code & generate python code and provide answer to the question \\ 
 & for completing the statement &  \\ 
\hline
Prompt with Context & Given context: \{context\} and the uncompleted  & Given context: \{context\} and the question: \{question\} \\
 & statement: \{statement\} \_ , provide answer and & generate explanation and provide answer to the question\\
 & generate explanation for completing the statement & \\ 
\hline
\end{tabular}
\vspace{1.0em}
\caption{Prompt Design}
\label{tb:prompt-design}
\vspace{-0.5em}
\end{table*}

\subsection{Metrics}
The main metric for measuring the performance of LLM reasoning is the accuracy of multi-hop reasoning conditioned on the correctness of each individual hop, denoted as $r=p(h|h_1, h_2, ..., h_n)$, where $h$ is the correctness of the multi-hop reasoning, and $h_i$ is the correctness of the $ith$ hop reasoning. We use this metric instead of the overall accuracy of the results because the final reasoning output is affected by multiple factors, including whether the LLM has the knowledge of each individual entity and whether it can infer each hop correctly. Our main metric measures the latent multi-hop reasoning performance. We also provide the overall accuracy of the results as a reference for potential applications.

\subsection{Experimental Results}
We compare the different approaches of grounding LLM reasoning using different knowledge graph representations by answering the following research questions.

\subsubsection{RQ1: Do different representations of entity relationships affect LLM multi-hop reasoning ? Can we fine-tune the LLM with proper entity relationship representations to improve its reasoning capabilities?}  

\noindent Since LLMs have already demonstrated abilities in latent multi-hop reasoning (\cite{yang2024largelanguagemodelslatently}), we designed experiments to compare how different approaches and different representations of multi-hop entity relationships can further improve LLMs' multi-hop reasoning ability. 

\textit{Impact of Entity Relationship Representations for LLM Prompting}. Table \ref{tb:dataset1_reasoning} and Table \ref{tb:dataset2_reasoning} present the performance of Llama-3.1-8B-Instruct with different representations across different datasets. It is obvious from the one-shot prompting results that Python representations of entity relationships outperform both the plain natural language text representation and the JSON representation. And both Python and natural language representations for one shot prompting perform better than zero shot prompting. The two-hop reasoning performance of the one-shot LLM with dynamic Python representation is approximately 78\% higher than that of zero-shot prompting of the LLM on the Dataset 1 (Table \ref{tb:dataset1_train_test_split}). Similarly, the performance of the static Python representation for one-shot prompting of the LLM is about 60\% higher than that of zero-shot prompting for Dataset 2 (Table \ref{tb:dataset2_train_test_split}). However, the performance of one-shot prompting with JSON representation is worse than zero-shot prompting of LLM. We hypothesize that the structured JSON representation is not native to the LLM reasoning process. 

\textit{Impact of Entity Relationship Representation for LLM Fine-Tuning}. Table \ref{tb:dataset1_reasoning} and Table \ref{tb:dataset2_reasoning} also show the performance of the fine-tuned Llama-3.1-8B-Instruct model with different representations on different datasets. In this case, the LLMs fine-tuned with Python representations perform better than the LLMs fine-tuned with either JSON representation or plain natural language text representations. The performance of the LLM fine-tuned with JSON representation is only slightly worse than that of the LLM fined-tuned with natural language representation. For comparison, we also provide the performance numbers for zero shot prompting of Llama-3.1-70B-Instruct. All LLMs that were fine-tuned with any entity relationship representation outperform the larger model for latent multi-hop reasoning.  

\begin{table*}[t]
\centering
\small
\begin{tabular}{llllcccc}
\hline
Model & KG Representation & KG Representation & Prompt & Accuracy & 1st hop correct & 1st hop correct & 1st hop correct \\
& for fine tuning & for Prompt & & & 2nd hop correct & 2nd hop correct & 2nd hop correct \\
& & & & & final incorrect & final correct & final accuracy \\
\hline
Llama-3.1-8B & NA & NA & Zero Shot & 16.3\% & 989 & 604 & 38.2\% \\
Llama-3.1-70B & NA & NA & Zero Shot  & 33.9\% & 615 & 1,979 & 76.3\% \\
Llama-3.1-8B & NA & Natural Language & One Shot & 16.9\% & 965 & 776 & 44.6\% \\
Llama-3.1-8B & NA & JSON & One Shot & 7.78\% & 995 & 352 & 26.1\% \\
Llama-3.1-8B & NA & Python Static & One Shot & 17.5\% & 408 & 609 & 59.9\% \\
Llama-3.1-8B & NA & Python Dynamic & One Shot & 19.1\% & 341 & 772 & 67.9\% \\
Llama-3.1-8B tuned & Natural Language & NA & Zero Shot & 25.6\% & 145 & 1,005 & 87.4\% \\
Llama-3.1-8B tuned & JSON & NA & Zero Shot & 20.8\% & 82 & 569 & 87.4\% \\
Llama-3.1-8B tuned & Python Static & NA & Zero Shot & 26.2\% & 102 & 1,093 & \textbf{91.5\%} \\
Llama-3.1-8B tuned & Python Dynamic & NA & Zero Shot & \textbf{26.5\%} & 119 & 882 & 88.1\% \\
\hline
\end{tabular}
\vspace{1em}
\caption{LLM Two Hop Reasoning for Dataset 1}
\label{tb:dataset1_reasoning}
\vspace{-0.5em}
\end{table*}

\begin{table*}[t]
\centering
\small
\begin{tabular}{llllcccc}
\hline
Model & KG Representation & KG Representation & Prompt & Accuracy & 1st hop correct & 1st hop correct & 1st hop correct \\
& for fine tuning & for Prompt & & & 2nd hop correct & 2nd hop correct & 2nd hop correct \\
& & & & & final incorrect & final correct & final accuracy \\
\hline
Llama-3.1-8B & NA & NA & Zero Shot & 1.85\% & 51 & 14 & 21.5\% \\
Llama-3.1-70B & NA & NA & Zero Shot & 10.7\% & 164 & 195 & 54.3\% \\
Llama-3.1-8B & NA & Natural Language & One Shot & 3.99\% & 82 & 25 & 23.4\% \\
Llama-3.1-8B & NA & JSON & One Shot & 1.80\% & 19 & 4 & 17.4\% \\
Llama-3.1-8B & NA & Python Static & One Shot & 4.98\% & 50 & 29 & 36.7\% \\
Llama-3.1-8B & NA & Python Dynamic & One Shot & 4.12\% & 35 & 12 & 25.5\% \\
Llama-3.1-8B tuned & Natural Language & NA & Zero Shot & 11.0\% & 97 & 164 & 62.8\% \\
Llama-3.1-8B tuned & JSON & NA & Zero Shot & 10.0\% & 61 & 157 & 72.0\% \\
Llama-3.1-8B tuned & Python Static & NA & Zero Shot & 12.1\% & 44 & 191 & 81.3\% \\
Llama-3.1-8B tuned & Python Dynamic & NA & Zero Shot & \textbf{12.3\%} & 38 & 203 & \textbf{84.2\%} \\
\hline
\end{tabular}
\vspace{1em}
\caption{LLM Two Hop Reasoning for Dataset 2}
\label{tb:dataset2_reasoning}
\vspace{-0.5em}
\end{table*}

\subsubsection{RQ2: Can fine-tuned LLMs generalize their reasoning ability to more hops that are beyond the training data ?}

\noindent Since we fine-tuned LLMs using only one-hop and two-hop reasoning data, it is important to study whether such a fine-tuning process will improve the LLMs' reasoning performance on more hops. We created a three-hop dataset (as shown in Table \ref{tb:dataset3_info}) to measure the performance of the fine-tuned LLMs. As shown in Table \ref{tb:dataset3_reasoning}, the three-hop reasoning performance of all fine-tuned LLMs has improved across all entity relationship representations compared with the baseline LLM without fine-tuning. Furthermore, LLMs that were fine-tuned with the Python representation outperform those fine-tuned with either the plain natural language representation or the JSON representation. As the relationship $(r_2, r_3)$ has significant overlap with $(r_1, r_2)$ in the training data (as shown in Table \ref{tb:dataset3_info}), the relative performance of fine-tuned models is consistent with what is shown in Table \ref{tb:dataset1_reasoning}. 

\begin{table*}[t]
\centering
\small
\begin{tabular}{lllllll}
\hline
Model & KG Representation & Accuracy & \% of Correct & \% of Correct & \% of Correct \\
& for tuning & & conditioned on & conditioned on & conditioned on \\
& & & 1st \& 2nd hop correct & 2nd \& 3rd hop correct & all three hops correct \\
\hline
Llama-3.1-8B & NA & 20.4\% & 35.2\% & 37.1\% & 50.0\% \\
Llama-3.1-70B & NA & 42.6\% & 64.6\% & 61.8\% & 80.4\% \\
Llama-3.1-8B tuned & Natural Language & 31.0\% & 44.0\% & 54.1\% & 65.0\% \\
Llama-3.1-8B tuned & JSON & 23.3\% & 42.5\% & 46.1\% & 63.9\% \\
Llama-3.1-8B tuned & Python Static & 30.9\% & 47.7\% & \textbf{59.3\%} & \textbf{70.6\%} \\
Llama-3.1-8B tuned & Python Dynamic & \textbf{38.0\%} & \textbf{51.6\%} & 57.4\% & 67.0\% \\
\hline
\end{tabular}
\vspace{1em}
\caption{LLM Three Hop Reasoning for Dataset 3 with Zero Shot Prompting}
\label{tb:dataset3_reasoning}
\vspace{-0.5em}
\end{table*}

\subsubsection{RQ3: How much can LLM in-context learning help or benefit from multi-hop reasoning?} 

\noindent Retrieval Augmented Generation (RAG) is one of the main applications of LLMs. However, even when supplied with correctly retrieved information, LLMs do not always generate the correct answer, particularly when multi-hop reasoning is required. To address this issue, we designed experiments to study the performance of fine-tuned LLMs when given an input context. In these experiments, we performed one-shot prompting of the LLMs, using the prompts listed in Table \ref{tb:prompt-design}. The results are shown in Table \ref{tb:context_reasoning}. For Dataset 1, since no context is provided, we use the 1st hop and 2nd hop inferences as context and measure whether the model can infer the correct answers. For Dataset 2, we use the given context for the questions. And again, the LLMs that were fine-tuned with different entity relationship representations outperform the baseline model without fine-tuning. Notably, the LLM fine-tuned with dynamic Python representation greatly outperforms the baseline model; it even surpasses the performance of the much larger baseline model (Llama-3.1-70B-Instruct).

\begin{table*}[t]
\centering
\small
\begin{tabular}{lllll}
\hline
Model & KG Representation & Dataset 1 Accuracy & Dataset 2 Accuracy \\
& for tuning & given 1st \& 2nd & given context \\
& & hop as context &   \\
\hline
Llama-3.1-8B & NA & 88.6\% & 10.9\% \\
Llama-3.1-70B & NA & 96.1\% & 41.9\% \\
Llama-3.1-8B tuned & Natural Language & 87.9\% & 44.7\% \\
Llama-3.1-8B tuned & JSON & 92.3\% & 46.6\% \\
Llama-3.1-8B tuned & Python Static & 87.2\% & 52.8\% \\
Llama-3.1-8B tuned & Python Dynamic & \textbf{96.4\%} & \textbf{59.2\%} \\
\hline
\end{tabular}
\vspace{1em}
\caption{LLM Two Hop Reasoning with Input Context}
\label{tb:context_reasoning}
\vspace{-0.5em}
\end{table*}

\subsection{Discussion} 
We designed a series of experiments to study how to seamlessly integrate entity relationships into LLMs to improve their multi-hop reasoning ability, and the impact of different entity relationship representations on the performance of LLMs. Our experimental results show that it is possible to integrate entity relationships into LLMs and ground the LLM multi-hop reasoning process with knowledge graphs. While all forms of the proposed entity relationship representations help improve LLM reasoning performance, they affect LLM performance differently. The natural language representation is straightforward and is the major form of pre-training data for LLMs. The JSON representation is best suited for storing structured data. However, it can be difficult for LLM to directly integrate pure structured data. The Python representations store both structured data and the inference process, providing a more controlled and unambiguous way of guiding LLMs through the reasoning process. This helps LLMs achieve better reasoning performance in all the cases we studied. In some cases, the fine-tuned small LLMs perform better than much larger LLMs, even though the deep neural networks of the larger LLMs can help the model make connections among multiple hops and infer the correct answers. Because we guide and fine-tune the models with emphasis on multi-hop relationships rather than the facts of individual entities, the fine-tuned models can easily generalize to do multi-hop reasoning for completely different entities, even in cases where the number of hops is greater than what is in the training data. Our proposed fine-tuning approaches also help improve the results of in-context learning.

As synthetic data becomes increasingly important for LLM pre-training and fine-tuning, generating proper representations of structured data (especially knowledge graphs) and incorporating this type of data in LLM pre-training and fine-tuning can greatly improve LLM reasoning abilities and reduce hallucinations. As demonstrated by our experimental results, the representation of entity relationships is very important for LLM performance. Programming languages provide the flexibility to represent various entity relationships with native data structures. They can guide LLM reasoning in a more controlled and principled way and ground LLM inference with a knowledge base.

In our experiments, we only studied the simplest form of reasoning: two-hop and three-hop reasoning with compositional relationships. The reasoning process for real-life applications can be much more sophisticated. As shown in Figure~\ref{fig:python-v2}, the Python representation of a knowledge graph with dynamic relationships provides the flexibility to define any relationships, even extending to a subgraph. Entities and relationships can be defined as classes themselves, allowing us to add attributes to the entities and relationships. Correspondingly, the ``infer'' function can be redefined to include code that checks these attributes. In future work, we will study the graph representation of entity relationships and its impact on more complex reasoning cases.

\section{Concluding Remarks} \label{conclusion}
We proposed different representations of entity relationships in knowledge graphs to improve the multi-hop reasoning capabilities of LLMs. We conducted a series of experiments to study how different representations of entity relationships affect LLM reasoning ability. We showed that introducing programming language representations of the entity relationships helps improve LLM multi-hop reasoning ability and reduce hallucination. 

The programming language representation of the entity relationships provides a controlled and unambiguous way for LLM multi-hop reasoning. By leveraging the native data structures inherent in programming languages, we can effectively model complex entity relationships, while iterative inference functions guide the logical reasoning process. This approach not only enhances reasoning accuracy but also facilitates generalization to more sophisticated reasoning use cases. However, accurately measuring the performance of LLM reasoning beyond two-hop and three-hop compositional relationships can be challenging, as the reasoning process becomes increasingly complex. As part of the future work, we would like to study the programming language representations of more sophisticated relationships in order to solve more complex reasoning tasks. We will experiment at both the pre-training and fine-tuning stages of LLMs to evaluate the performance impact. We hope that our work will inspire other researchers to further push the frontier of LLM research.

\bibliographystyle{ACM-Reference-Format}
\bibliography{reference}


\begin{thebibliography}{40}


\ifx \showCODEN    \undefined \def \showCODEN     #1{\unskip}     \fi
\ifx \showDOI      \undefined \def \showDOI       #1{#1}\fi
\ifx \showISBNx    \undefined \def \showISBNx     #1{\unskip}     \fi
\ifx \showISBNxiii \undefined \def \showISBNxiii  #1{\unskip}     \fi
\ifx \showISSN     \undefined \def \showISSN      #1{\unskip}     \fi
\ifx \showLCCN     \undefined \def \showLCCN      #1{\unskip}     \fi
\ifx \shownote     \undefined \def \shownote      #1{#1}          \fi
\ifx \showarticletitle \undefined \def \showarticletitle #1{#1}   \fi
\ifx \showURL      \undefined \def \showURL       {\relax}        \fi
\providecommand\bibfield[2]{#2}
\providecommand\bibinfo[2]{#2}
\providecommand\natexlab[1]{#1}
\providecommand\showeprint[2][]{arXiv:#2}

\bibitem[AI(2024)]%
        {dubey2024llama3herdmodels}
\bibfield{author}{\bibinfo{person}{Meta AI}.} \bibinfo{year}{2024}\natexlab{}.
\newblock \bibinfo{title}{The Llama 3 Herd of Models}.
\newblock
\newblock
\showeprint[arxiv]{2407.21783}~[cs.AI]
\urldef\tempurl%
\url{https://arxiv.org/abs/2407.21783}
\showURL{%
\tempurl}


\bibitem[Aryabumi et~al\mbox{.}(2024)]%
        {aryabumi2024codecodeexploringimpact}
\bibfield{author}{\bibinfo{person}{Viraat Aryabumi}, \bibinfo{person}{Yixuan
  Su}, \bibinfo{person}{Raymond Ma}, \bibinfo{person}{Adrien Morisot},
  \bibinfo{person}{Ivan Zhang}, \bibinfo{person}{Acyr Locatelli},
  \bibinfo{person}{Marzieh Fadaee}, \bibinfo{person}{Ahmet Üstün}, {and}
  \bibinfo{person}{Sara Hooker}.} \bibinfo{year}{2024}\natexlab{}.
\newblock \bibinfo{title}{To Code, or Not To Code? Exploring Impact of Code in
  Pre-training}.
\newblock
\newblock
\showeprint[arxiv]{2408.10914}~[cs.CL]
\urldef\tempurl%
\url{https://arxiv.org/abs/2408.10914}
\showURL{%
\tempurl}


\bibitem[Biran et~al\mbox{.}(2024)]%
        {biran2024hoppinglateexploringlimitations}
\bibfield{author}{\bibinfo{person}{Eden Biran}, \bibinfo{person}{Daniela
  Gottesman}, \bibinfo{person}{Sohee Yang}, \bibinfo{person}{Mor Geva}, {and}
  \bibinfo{person}{Amir Globerson}.} \bibinfo{year}{2024}\natexlab{}.
\newblock \bibinfo{title}{Hopping Too Late: Exploring the Limitations of Large
  Language Models on Multi-Hop Queries}.
\newblock
\newblock
\showeprint[arxiv]{2406.12775}~[cs.CL]
\urldef\tempurl%
\url{https://arxiv.org/abs/2406.12775}
\showURL{%
\tempurl}


\bibitem[Bollacker et~al\mbox{.}(2008)]%
        {Bollacker2008FreebaseAC}
\bibfield{author}{\bibinfo{person}{Kurt~D. Bollacker}, \bibinfo{person}{Colin
  Evans}, \bibinfo{person}{Praveen~K. Paritosh}, \bibinfo{person}{Tim Sturge},
  {and} \bibinfo{person}{Jamie Taylor}.} \bibinfo{year}{2008}\natexlab{}.
\newblock \showarticletitle{Freebase: a collaboratively created graph database
  for structuring human knowledge}. In \bibinfo{booktitle}{\emph{SIGMOD
  Conference}}.
\newblock
\urldef\tempurl%
\url{https://api.semanticscholar.org/CorpusID:207167677}
\showURL{%
\tempurl}


\bibitem[Bray(2014)]%
        {bray2014javascript}
\bibfield{author}{\bibinfo{person}{Tim Bray}.} \bibinfo{year}{2014}\natexlab{}.
\newblock \bibinfo{title}{The javascript object notation (json) data
  interchange format}.
\newblock
\newblock


\bibitem[Brei et~al\mbox{.}(2024)]%
        {brei2024leveragingsmalllanguagemodels}
\bibfield{author}{\bibinfo{person}{Felix Brei}, \bibinfo{person}{Johannes
  Frey}, {and} \bibinfo{person}{Lars-Peter Meyer}.}
  \bibinfo{year}{2024}\natexlab{}.
\newblock \bibinfo{title}{Leveraging small language models for Text2SPARQL
  tasks to improve the resilience of AI assistance}.
\newblock
\newblock
\showeprint[arxiv]{2405.17076}~[cs.AI]
\urldef\tempurl%
\url{https://arxiv.org/abs/2405.17076}
\showURL{%
\tempurl}


\bibitem[Brown et~al\mbox{.}(2020)]%
        {brown2020languagemodelsfewshotlearners}
\bibfield{author}{\bibinfo{person}{Tom~B. Brown}, \bibinfo{person}{Benjamin
  Mann}, \bibinfo{person}{Nick Ryder}, \bibinfo{person}{Melanie Subbiah},
  \bibinfo{person}{Jared Kaplan}, \bibinfo{person}{Prafulla Dhariwal},
  \bibinfo{person}{Arvind Neelakantan}, \bibinfo{person}{Pranav Shyam},
  \bibinfo{person}{Girish Sastry}, \bibinfo{person}{Amanda Askell},
  \bibinfo{person}{Sandhini Agarwal}, \bibinfo{person}{Ariel Herbert-Voss},
  \bibinfo{person}{Gretchen Krueger}, \bibinfo{person}{Tom Henighan},
  \bibinfo{person}{Rewon Child}, \bibinfo{person}{Aditya Ramesh},
  \bibinfo{person}{Daniel~M. Ziegler}, \bibinfo{person}{Jeffrey Wu},
  \bibinfo{person}{Clemens Winter}, \bibinfo{person}{Christopher Hesse},
  \bibinfo{person}{Mark Chen}, \bibinfo{person}{Eric Sigler},
  \bibinfo{person}{Mateusz Litwin}, \bibinfo{person}{Scott Gray},
  \bibinfo{person}{Benjamin Chess}, \bibinfo{person}{Jack Clark},
  \bibinfo{person}{Christopher Berner}, \bibinfo{person}{Sam McCandlish},
  \bibinfo{person}{Alec Radford}, \bibinfo{person}{Ilya Sutskever}, {and}
  \bibinfo{person}{Dario Amodei}.} \bibinfo{year}{2020}\natexlab{}.
\newblock \bibinfo{title}{Language Models are Few-Shot Learners}.
\newblock
\newblock
\showeprint[arxiv]{2005.14165}~[cs.CL]
\urldef\tempurl%
\url{https://arxiv.org/abs/2005.14165}
\showURL{%
\tempurl}


\bibitem[Bustamante and Takeda(2024)]%
        {bustamante2024sparqlgenerationentitypretrained}
\bibfield{author}{\bibinfo{person}{Diego Bustamante} {and}
  \bibinfo{person}{Hideaki Takeda}.} \bibinfo{year}{2024}\natexlab{}.
\newblock \bibinfo{title}{SPARQL Generation with Entity Pre-trained GPT for KG
  Question Answering}.
\newblock
\newblock
\showeprint[arxiv]{2402.00969}~[cs.CL]
\urldef\tempurl%
\url{https://arxiv.org/abs/2402.00969}
\showURL{%
\tempurl}


\bibitem[Chai et~al\mbox{.}(2023)]%
        {chai2023graphllmboostinggraphreasoning}
\bibfield{author}{\bibinfo{person}{Ziwei Chai}, \bibinfo{person}{Tianjie
  Zhang}, \bibinfo{person}{Liang Wu}, \bibinfo{person}{Kaiqiao Han},
  \bibinfo{person}{Xiaohai Hu}, \bibinfo{person}{Xuanwen Huang}, {and}
  \bibinfo{person}{Yang Yang}.} \bibinfo{year}{2023}\natexlab{}.
\newblock \bibinfo{title}{GraphLLM: Boosting Graph Reasoning Ability of Large
  Language Model}.
\newblock
\newblock
\showeprint[arxiv]{2310.05845}~[cs.CL]
\urldef\tempurl%
\url{https://arxiv.org/abs/2310.05845}
\showURL{%
\tempurl}


\bibitem[Choudhary et~al\mbox{.}(2022)]%
        {Choudhary2022}
\bibfield{author}{\bibinfo{person}{Nurendra Choudhary}, \bibinfo{person}{Nikhil
  Rao}, \bibinfo{person}{Karthik Subbian}, {and} \bibinfo{person}{Chandan
  Reddy}.} \bibinfo{year}{2022}\natexlab{}.
\newblock \showarticletitle{Graph-based multilingual language model: Leveraging
  product relations for search relevance}. In \bibinfo{booktitle}{\emph{KDD
  2022}}.
\newblock
\urldef\tempurl%
\url{https://www.amazon.science/publications/graph-based-multilingual-language-model-leveraging-product-relations-for-search-relevance}
\showURL{%
\tempurl}


\bibitem[Chowdhery et~al\mbox{.}(2022)]%
        {chowdhery2022palmscalinglanguagemodeling}
\bibfield{author}{\bibinfo{person}{Aakanksha Chowdhery},
  \bibinfo{person}{Sharan Narang}, \bibinfo{person}{Jacob Devlin},
  \bibinfo{person}{Maarten Bosma}, \bibinfo{person}{Gaurav Mishra},
  \bibinfo{person}{Adam Roberts}, \bibinfo{person}{Paul Barham},
  \bibinfo{person}{Hyung~Won Chung}, \bibinfo{person}{Charles Sutton},
  \bibinfo{person}{Sebastian Gehrmann}, \bibinfo{person}{Parker Schuh},
  \bibinfo{person}{Kensen Shi}, \bibinfo{person}{Sasha Tsvyashchenko},
  \bibinfo{person}{Joshua Maynez}, \bibinfo{person}{Abhishek Rao},
  \bibinfo{person}{Parker Barnes}, \bibinfo{person}{Yi Tay},
  \bibinfo{person}{Noam Shazeer}, \bibinfo{person}{Vinodkumar Prabhakaran},
  \bibinfo{person}{Emily Reif}, \bibinfo{person}{Nan Du}, \bibinfo{person}{Ben
  Hutchinson}, \bibinfo{person}{Reiner Pope}, \bibinfo{person}{James Bradbury},
  \bibinfo{person}{Jacob Austin}, \bibinfo{person}{Michael Isard},
  \bibinfo{person}{Guy Gur-Ari}, \bibinfo{person}{Pengcheng Yin},
  \bibinfo{person}{Toju Duke}, \bibinfo{person}{Anselm Levskaya},
  \bibinfo{person}{Sanjay Ghemawat}, \bibinfo{person}{Sunipa Dev},
  \bibinfo{person}{Henryk Michalewski}, \bibinfo{person}{Xavier Garcia},
  \bibinfo{person}{Vedant Misra}, \bibinfo{person}{Kevin Robinson},
  \bibinfo{person}{Liam Fedus}, \bibinfo{person}{Denny Zhou},
  \bibinfo{person}{Daphne Ippolito}, \bibinfo{person}{David Luan},
  \bibinfo{person}{Hyeontaek Lim}, \bibinfo{person}{Barret Zoph},
  \bibinfo{person}{Alexander Spiridonov}, \bibinfo{person}{Ryan Sepassi},
  \bibinfo{person}{David Dohan}, \bibinfo{person}{Shivani Agrawal},
  \bibinfo{person}{Mark Omernick}, \bibinfo{person}{Andrew~M. Dai},
  \bibinfo{person}{Thanumalayan~Sankaranarayana Pillai}, \bibinfo{person}{Marie
  Pellat}, \bibinfo{person}{Aitor Lewkowycz}, \bibinfo{person}{Erica Moreira},
  \bibinfo{person}{Rewon Child}, \bibinfo{person}{Oleksandr Polozov},
  \bibinfo{person}{Katherine Lee}, \bibinfo{person}{Zongwei Zhou},
  \bibinfo{person}{Xuezhi Wang}, \bibinfo{person}{Brennan Saeta},
  \bibinfo{person}{Mark Diaz}, \bibinfo{person}{Orhan Firat},
  \bibinfo{person}{Michele Catasta}, \bibinfo{person}{Jason Wei},
  \bibinfo{person}{Kathy Meier-Hellstern}, \bibinfo{person}{Douglas Eck},
  \bibinfo{person}{Jeff Dean}, \bibinfo{person}{Slav Petrov}, {and}
  \bibinfo{person}{Noah Fiedel}.} \bibinfo{year}{2022}\natexlab{}.
\newblock \bibinfo{title}{PaLM: Scaling Language Modeling with Pathways}.
\newblock
\newblock
\showeprint[arxiv]{2204.02311}~[cs.CL]
\urldef\tempurl%
\url{https://arxiv.org/abs/2204.02311}
\showURL{%
\tempurl}


\bibitem[Dernbach et~al\mbox{.}(2024)]%
        {dernbach2024glamfinetuninglargelanguage}
\bibfield{author}{\bibinfo{person}{Stefan Dernbach}, \bibinfo{person}{Khushbu
  Agarwal}, \bibinfo{person}{Alejandro Zuniga}, \bibinfo{person}{Michael
  Henry}, {and} \bibinfo{person}{Sutanay Choudhury}.}
  \bibinfo{year}{2024}\natexlab{}.
\newblock \bibinfo{title}{GLaM: Fine-Tuning Large Language Models for Domain
  Knowledge Graph Alignment via Neighborhood Partitioning and Generative
  Subgraph Encoding}.
\newblock
\newblock
\showeprint[arxiv]{2402.06764}~[cs.AI]
\urldef\tempurl%
\url{https://arxiv.org/abs/2402.06764}
\showURL{%
\tempurl}


\bibitem[Edge et~al\mbox{.}(2024)]%
        {edge2024localglobalgraphrag}
\bibfield{author}{\bibinfo{person}{Darren Edge}, \bibinfo{person}{Ha Trinh},
  \bibinfo{person}{Newman Cheng}, \bibinfo{person}{Joshua Bradley},
  \bibinfo{person}{Alex Chao}, \bibinfo{person}{Apurva Mody},
  \bibinfo{person}{Steven Truitt}, {and} \bibinfo{person}{Jonathan Larson}.}
  \bibinfo{year}{2024}\natexlab{}.
\newblock \bibinfo{title}{From Local to Global: A Graph RAG Approach to
  Query-Focused Summarization}.
\newblock
\newblock
\showeprint[arxiv]{2404.16130}~[cs.CL]
\urldef\tempurl%
\url{https://arxiv.org/abs/2404.16130}
\showURL{%
\tempurl}


\bibitem[Fu et~al\mbox{.}(2023)]%
        {fu2023complexitybased}
\bibfield{author}{\bibinfo{person}{Yao Fu}, \bibinfo{person}{Hao Peng},
  \bibinfo{person}{Ashish Sabharwal}, \bibinfo{person}{Peter Clark}, {and}
  \bibinfo{person}{Tushar Khot}.} \bibinfo{year}{2023}\natexlab{}.
\newblock \showarticletitle{Complexity-Based Prompting for Multi-step
  Reasoning}. In \bibinfo{booktitle}{\emph{The Eleventh International
  Conference on Learning Representations}}.
\newblock
\urldef\tempurl%
\url{https://openreview.net/forum?id=yf1icZHC-l9}
\showURL{%
\tempurl}


\bibitem[Guo et~al\mbox{.}(2023)]%
        {guo2023gpt4graphlargelanguagemodels}
\bibfield{author}{\bibinfo{person}{Jiayan Guo}, \bibinfo{person}{Lun Du},
  \bibinfo{person}{Hengyu Liu}, \bibinfo{person}{Mengyu Zhou},
  \bibinfo{person}{Xinyi He}, {and} \bibinfo{person}{Shi Han}.}
  \bibinfo{year}{2023}\natexlab{}.
\newblock \bibinfo{title}{GPT4Graph: Can Large Language Models Understand Graph
  Structured Data ? An Empirical Evaluation and Benchmarking}.
\newblock
\newblock
\showeprint[arxiv]{2305.15066}~[cs.AI]
\urldef\tempurl%
\url{https://arxiv.org/abs/2305.15066}
\showURL{%
\tempurl}


\bibitem[Ho et~al\mbox{.}(2020)]%
        {xanh2020_2wikimultihop}
\bibfield{author}{\bibinfo{person}{Xanh Ho}, \bibinfo{person}{Anh-Khoa
  Duong~Nguyen}, \bibinfo{person}{Saku Sugawara}, {and} \bibinfo{person}{Akiko
  Aizawa}.} \bibinfo{year}{2020}\natexlab{}.
\newblock \showarticletitle{Constructing A Multi-hop {QA} Dataset for
  Comprehensive Evaluation of Reasoning Steps}. In
  \bibinfo{booktitle}{\emph{Proceedings of the 28th International Conference on
  Computational Linguistics}}. \bibinfo{publisher}{International Committee on
  Computational Linguistics}, \bibinfo{address}{Barcelona, Spain (Online)},
  \bibinfo{pages}{6609--6625}.
\newblock
\urldef\tempurl%
\url{https://www.aclweb.org/anthology/2020.coling-main.580}
\showURL{%
\tempurl}


\bibitem[Hu et~al\mbox{.}(2022)]%
        {hu2022lora}
\bibfield{author}{\bibinfo{person}{Edward~J Hu}, \bibinfo{person}{Yelong Shen},
  \bibinfo{person}{Phillip Wallis}, \bibinfo{person}{Zeyuan Allen-Zhu},
  \bibinfo{person}{Yuanzhi Li}, \bibinfo{person}{Shean Wang},
  \bibinfo{person}{Lu Wang}, {and} \bibinfo{person}{Weizhu Chen}.}
  \bibinfo{year}{2022}\natexlab{}.
\newblock \showarticletitle{Lo{RA}: Low-Rank Adaptation of Large Language
  Models}. In \bibinfo{booktitle}{\emph{International Conference on Learning
  Representations}}.
\newblock
\urldef\tempurl%
\url{https://openreview.net/forum?id=nZeVKeeFYf9}
\showURL{%
\tempurl}


\bibitem[Ji et~al\mbox{.}(2023)]%
        {Ji_2023}
\bibfield{author}{\bibinfo{person}{Ziwei Ji}, \bibinfo{person}{Nayeon Lee},
  \bibinfo{person}{Rita Frieske}, \bibinfo{person}{Tiezheng Yu},
  \bibinfo{person}{Dan Su}, \bibinfo{person}{Yan Xu}, \bibinfo{person}{Etsuko
  Ishii}, \bibinfo{person}{Ye~Jin Bang}, \bibinfo{person}{Andrea Madotto},
  {and} \bibinfo{person}{Pascale Fung}.} \bibinfo{year}{2023}\natexlab{}.
\newblock \showarticletitle{Survey of Hallucination in Natural Language
  Generation}.
\newblock \bibinfo{journal}{\emph{Comput. Surveys}} \bibinfo{volume}{55},
  \bibinfo{number}{12} (\bibinfo{date}{March} \bibinfo{year}{2023}),
  \bibinfo{pages}{1–38}.
\newblock
\showISSN{1557-7341}
\urldef\tempurl%
\url{https://doi.org/10.1145/3571730}
\showDOI{\tempurl}


\bibitem[Kim et~al\mbox{.}(2023)]%
        {kim2023cotcollectionimprovingzeroshot}
\bibfield{author}{\bibinfo{person}{Seungone Kim}, \bibinfo{person}{Se~June
  Joo}, \bibinfo{person}{Doyoung Kim}, \bibinfo{person}{Joel Jang},
  \bibinfo{person}{Seonghyeon Ye}, \bibinfo{person}{Jamin Shin}, {and}
  \bibinfo{person}{Minjoon Seo}.} \bibinfo{year}{2023}\natexlab{}.
\newblock \bibinfo{title}{The CoT Collection: Improving Zero-shot and Few-shot
  Learning of Language Models via Chain-of-Thought Fine-Tuning}.
\newblock
\newblock
\showeprint[arxiv]{2305.14045}~[cs.CL]
\urldef\tempurl%
\url{https://arxiv.org/abs/2305.14045}
\showURL{%
\tempurl}


\bibitem[Lazaridou et~al\mbox{.}(2022)]%
        {lazaridou2022internetaugmentedlanguagemodelsfewshot}
\bibfield{author}{\bibinfo{person}{Angeliki Lazaridou}, \bibinfo{person}{Elena
  Gribovskaya}, \bibinfo{person}{Wojciech Stokowiec}, {and}
  \bibinfo{person}{Nikolai Grigorev}.} \bibinfo{year}{2022}\natexlab{}.
\newblock \bibinfo{title}{Internet-augmented language models through few-shot
  prompting for open-domain question answering}.
\newblock
\newblock
\showeprint[arxiv]{2203.05115}~[cs.CL]
\urldef\tempurl%
\url{https://arxiv.org/abs/2203.05115}
\showURL{%
\tempurl}


\bibitem[Lehmann et~al\mbox{.}(2014)]%
        {Lehmann2014DBpedia}
\bibfield{author}{\bibinfo{person}{Jens Lehmann}, \bibinfo{person}{Robert
  Isele}, \bibinfo{person}{Max Jakob}, \bibinfo{person}{Anja Jentzsch},
  \bibinfo{person}{Dimitris Kontokostas}, \bibinfo{person}{Pablo Mendes},
  \bibinfo{person}{Sebastian Hellmann}, \bibinfo{person}{Mohamed Morsey},
  \bibinfo{person}{Patrick Van~Kleef}, \bibinfo{person}{Sören Auer}, {and}
  \bibinfo{person}{Christian Bizer}.} \bibinfo{year}{2014}\natexlab{}.
\newblock \showarticletitle{DBpedia - A Large-scale, Multilingual Knowledge
  Base Extracted from Wikipedia}.
\newblock \bibinfo{journal}{\emph{Semantic Web Journal}}  \bibinfo{volume}{6}
  (\bibinfo{date}{01} \bibinfo{year}{2014}).
\newblock
\urldef\tempurl%
\url{https://doi.org/10.3233/SW-140134}
\showDOI{\tempurl}


\bibitem[Luo et~al\mbox{.}(2024)]%
        {luo2024reasoninggraphsfaithfulinterpretable}
\bibfield{author}{\bibinfo{person}{Linhao Luo}, \bibinfo{person}{Yuan-Fang Li},
  \bibinfo{person}{Gholamreza Haffari}, {and} \bibinfo{person}{Shirui Pan}.}
  \bibinfo{year}{2024}\natexlab{}.
\newblock \bibinfo{title}{Reasoning on Graphs: Faithful and Interpretable Large
  Language Model Reasoning}.
\newblock
\newblock
\showeprint[arxiv]{2310.01061}~[cs.CL]
\urldef\tempurl%
\url{https://arxiv.org/abs/2310.01061}
\showURL{%
\tempurl}


\bibitem[Ma et~al\mbox{.}(2023)]%
        {ma2023trainingstagedoescode}
\bibfield{author}{\bibinfo{person}{Yingwei Ma}, \bibinfo{person}{Yue Liu},
  \bibinfo{person}{Yue Yu}, \bibinfo{person}{Yuanliang Zhang},
  \bibinfo{person}{Yu Jiang}, \bibinfo{person}{Changjian Wang}, {and}
  \bibinfo{person}{Shanshan Li}.} \bibinfo{year}{2023}\natexlab{}.
\newblock \bibinfo{title}{At Which Training Stage Does Code Data Help LLMs
  Reasoning?}
\newblock
\newblock
\showeprint[arxiv]{2309.16298}~[cs.CL]
\urldef\tempurl%
\url{https://arxiv.org/abs/2309.16298}
\showURL{%
\tempurl}


\bibitem[Nie et~al\mbox{.}(2024)]%
        {nie2024codestyleincontextlearningknowledgebased}
\bibfield{author}{\bibinfo{person}{Zhijie Nie}, \bibinfo{person}{Richong
  Zhang}, \bibinfo{person}{Zhongyuan Wang}, {and} \bibinfo{person}{Xudong
  Liu}.} \bibinfo{year}{2024}\natexlab{}.
\newblock \bibinfo{title}{Code-Style In-Context Learning for Knowledge-Based
  Question Answering}.
\newblock
\newblock
\showeprint[arxiv]{2309.04695}~[cs.CL]
\urldef\tempurl%
\url{https://arxiv.org/abs/2309.04695}
\showURL{%
\tempurl}


\bibitem[Press et~al\mbox{.}(2023)]%
        {pressMeasuringNarrowingCompositionality2023}
\bibfield{author}{\bibinfo{person}{Ofir Press}, \bibinfo{person}{Muru Zhang},
  \bibinfo{person}{Sewon Min}, \bibinfo{person}{Ludwig Schmidt},
  \bibinfo{person}{Noah Smith}, {and} \bibinfo{person}{Mike Lewis}.}
  \bibinfo{year}{2023}\natexlab{}.
\newblock \showarticletitle{Measuring and {{Narrowing}} the {{Compositionality
  Gap}} in {{Language Models}}}. In \bibinfo{booktitle}{\emph{Findings of the
  {{Association}} for {{Computational Linguistics}}: {{EMNLP}} 2023}},
  \bibfield{editor}{\bibinfo{person}{Houda Bouamor}, \bibinfo{person}{Juan
  Pino}, {and} \bibinfo{person}{Kalika Bali}} (Eds.).
  \bibinfo{publisher}{Association for Computational Linguistics},
  \bibinfo{address}{Singapore}, \bibinfo{pages}{5687--5711}.
\newblock
\urldef\tempurl%
\url{https://doi.org/10.18653/v1/2023.findings-emnlp.378}
\showDOI{\tempurl}


\bibitem[Rangel et~al\mbox{.}(2024)]%
        {rangel2024sparqlgenerationanalysisfinetuning}
\bibfield{author}{\bibinfo{person}{Julio~C. Rangel},
  \bibinfo{person}{Tarcisio~Mendes de Farias}, \bibinfo{person}{Ana~Claudia
  Sima}, {and} \bibinfo{person}{Norio Kobayashi}.}
  \bibinfo{year}{2024}\natexlab{}.
\newblock \bibinfo{title}{SPARQL Generation: an analysis on fine-tuning
  OpenLLaMA for Question Answering over a Life Science Knowledge Graph}.
\newblock
\newblock
\showeprint[arxiv]{2402.04627}~[cs.AI]
\urldef\tempurl%
\url{https://arxiv.org/abs/2402.04627}
\showURL{%
\tempurl}


\bibitem[Vrande\v{c}i\'{c} and Kr\"{o}tzsch(2014)]%
        {vrandevcic2014wikidata}
\bibfield{author}{\bibinfo{person}{Denny Vrande\v{c}i\'{c}} {and}
  \bibinfo{person}{Markus Kr\"{o}tzsch}.} \bibinfo{year}{2014}\natexlab{}.
\newblock \showarticletitle{Wikidata: A Free Collaborative Knowledgebase}.
\newblock \bibinfo{journal}{\emph{Commun. ACM}} \bibinfo{volume}{57},
  \bibinfo{number}{10} (\bibinfo{date}{Sept.} \bibinfo{year}{2014}),
  \bibinfo{pages}{78--85}.
\newblock
\showISSN{0001-0782}
\urldef\tempurl%
\url{https://doi.org/10.1145/2629489}
\showDOI{\tempurl}


\bibitem[Wang et~al\mbox{.}(2024a)]%
        {wang2024languagemodelssolvegraph}
\bibfield{author}{\bibinfo{person}{Heng Wang}, \bibinfo{person}{Shangbin Feng},
  \bibinfo{person}{Tianxing He}, \bibinfo{person}{Zhaoxuan Tan},
  \bibinfo{person}{Xiaochuang Han}, {and} \bibinfo{person}{Yulia Tsvetkov}.}
  \bibinfo{year}{2024}\natexlab{a}.
\newblock \bibinfo{title}{Can Language Models Solve Graph Problems in Natural
  Language?}
\newblock
\newblock
\showeprint[arxiv]{2305.10037}~[cs.CL]
\urldef\tempurl%
\url{https://arxiv.org/abs/2305.10037}
\showURL{%
\tempurl}


\bibitem[Wang et~al\mbox{.}(2024b)]%
        {wang2024instructgraphboostinglargelanguage}
\bibfield{author}{\bibinfo{person}{Jianing Wang}, \bibinfo{person}{Junda Wu},
  \bibinfo{person}{Yupeng Hou}, \bibinfo{person}{Yao Liu},
  \bibinfo{person}{Ming Gao}, {and} \bibinfo{person}{Julian McAuley}.}
  \bibinfo{year}{2024}\natexlab{b}.
\newblock \bibinfo{title}{InstructGraph: Boosting Large Language Models via
  Graph-centric Instruction Tuning and Preference Alignment}.
\newblock
\newblock
\showeprint[arxiv]{2402.08785}~[cs.CL]
\urldef\tempurl%
\url{https://arxiv.org/abs/2402.08785}
\showURL{%
\tempurl}


\bibitem[Wang et~al\mbox{.}(2023)]%
        {wang2023selfconsistency}
\bibfield{author}{\bibinfo{person}{Xuezhi Wang}, \bibinfo{person}{Jason Wei},
  \bibinfo{person}{Dale Schuurmans}, \bibinfo{person}{Quoc~V Le},
  \bibinfo{person}{Ed~H. Chi}, \bibinfo{person}{Sharan Narang},
  \bibinfo{person}{Aakanksha Chowdhery}, {and} \bibinfo{person}{Denny Zhou}.}
  \bibinfo{year}{2023}\natexlab{}.
\newblock \showarticletitle{Self-Consistency Improves Chain of Thought
  Reasoning in Language Models}. In \bibinfo{booktitle}{\emph{The Eleventh
  International Conference on Learning Representations}}.
\newblock
\urldef\tempurl%
\url{https://openreview.net/forum?id=1PL1NIMMrw}
\showURL{%
\tempurl}


\bibitem[Wei et~al\mbox{.}(2022)]%
        {wei2022chainofthought}
\bibfield{author}{\bibinfo{person}{Jason Wei}, \bibinfo{person}{Xuezhi Wang},
  \bibinfo{person}{Dale Schuurmans}, \bibinfo{person}{Maarten Bosma},
  \bibinfo{person}{brian ichter}, \bibinfo{person}{Fei Xia},
  \bibinfo{person}{Ed Chi}, \bibinfo{person}{Quoc~V Le}, {and}
  \bibinfo{person}{Denny Zhou}.} \bibinfo{year}{2022}\natexlab{}.
\newblock \showarticletitle{Chain-of-Thought Prompting Elicits Reasoning in
  Large Language Models}. In \bibinfo{booktitle}{\emph{Advances in Neural
  Information Processing Systems}},
  \bibfield{editor}{\bibinfo{person}{S.~Koyejo}, \bibinfo{person}{S.~Mohamed},
  \bibinfo{person}{A.~Agarwal}, \bibinfo{person}{D.~Belgrave},
  \bibinfo{person}{K.~Cho}, {and} \bibinfo{person}{A.~Oh}} (Eds.),
  Vol.~\bibinfo{volume}{35}. \bibinfo{publisher}{Curran Associates, Inc.},
  \bibinfo{pages}{24824--24837}.
\newblock
\urldef\tempurl%
\url{https://proceedings.neurips.cc/paper_files/paper/2022/file/9d5609613524ecf4f15af0f7b31abca4-Paper-Conference.pdf}
\showURL{%
\tempurl}


\bibitem[Xu et~al\mbox{.}(2023)]%
        {xu-etal-2023-fine}
\bibfield{author}{\bibinfo{person}{Silei Xu}, \bibinfo{person}{Shicheng Liu},
  \bibinfo{person}{Theo Culhane}, \bibinfo{person}{Elizaveta Pertseva},
  \bibinfo{person}{Meng-Hsi Wu}, \bibinfo{person}{Sina Semnani}, {and}
  \bibinfo{person}{Monica Lam}.} \bibinfo{year}{2023}\natexlab{}.
\newblock \showarticletitle{Fine-tuned {LLM}s Know More, Hallucinate Less with
  Few-Shot Sequence-to-Sequence Semantic Parsing over {W}ikidata}. In
  \bibinfo{booktitle}{\emph{Proceedings of the 2023 Conference on Empirical
  Methods in Natural Language Processing}},
  \bibfield{editor}{\bibinfo{person}{Houda Bouamor}, \bibinfo{person}{Juan
  Pino}, {and} \bibinfo{person}{Kalika Bali}} (Eds.).
  \bibinfo{publisher}{Association for Computational Linguistics},
  \bibinfo{address}{Singapore}, \bibinfo{pages}{5778--5791}.
\newblock
\urldef\tempurl%
\url{https://doi.org/10.18653/v1/2023.emnlp-main.353}
\showDOI{\tempurl}


\bibitem[Yang et~al\mbox{.}(2024)]%
        {yang2024largelanguagemodelslatently}
\bibfield{author}{\bibinfo{person}{Sohee Yang}, \bibinfo{person}{Elena
  Gribovskaya}, \bibinfo{person}{Nora Kassner}, \bibinfo{person}{Mor Geva},
  {and} \bibinfo{person}{Sebastian Riedel}.} \bibinfo{year}{2024}\natexlab{}.
\newblock \bibinfo{title}{Do Large Language Models Latently Perform Multi-Hop
  Reasoning?}
\newblock
\newblock
\showeprint[arxiv]{2402.16837}~[cs.CL]
\urldef\tempurl%
\url{https://arxiv.org/abs/2402.16837}
\showURL{%
\tempurl}


\bibitem[Yao et~al\mbox{.}(2023a)]%
        {yao2023tree}
\bibfield{author}{\bibinfo{person}{Shunyu Yao}, \bibinfo{person}{Dian Yu},
  \bibinfo{person}{Jeffrey Zhao}, \bibinfo{person}{Izhak Shafran},
  \bibinfo{person}{Thomas~L. Griffiths}, \bibinfo{person}{Yuan Cao}, {and}
  \bibinfo{person}{Karthik~R Narasimhan}.} \bibinfo{year}{2023}\natexlab{a}.
\newblock \showarticletitle{Tree of Thoughts: Deliberate Problem Solving with
  Large Language Models}. In \bibinfo{booktitle}{\emph{Thirty-seventh
  Conference on Neural Information Processing Systems}}.
\newblock
\urldef\tempurl%
\url{https://openreview.net/forum?id=5Xc1ecxO1h}
\showURL{%
\tempurl}


\bibitem[Yao et~al\mbox{.}(2023b)]%
        {yao2023react}
\bibfield{author}{\bibinfo{person}{Shunyu Yao}, \bibinfo{person}{Jeffrey Zhao},
  \bibinfo{person}{Dian Yu}, \bibinfo{person}{Nan Du}, \bibinfo{person}{Izhak
  Shafran}, \bibinfo{person}{Karthik Narasimhan}, {and} \bibinfo{person}{Yuan
  Cao}.} \bibinfo{year}{2023}\natexlab{b}.
\newblock \showarticletitle{{ReAct}: Synergizing Reasoning and Acting in
  Language Models}. In \bibinfo{booktitle}{\emph{International Conference on
  Learning Representations (ICLR)}}.
\newblock


\bibitem[Yasunaga et~al\mbox{.}(2022)]%
        {yasunaga2022deepbidirectionallanguageknowledgegraph}
\bibfield{author}{\bibinfo{person}{Michihiro Yasunaga},
  \bibinfo{person}{Antoine Bosselut}, \bibinfo{person}{Hongyu Ren},
  \bibinfo{person}{Xikun Zhang}, \bibinfo{person}{Christopher~D Manning},
  \bibinfo{person}{Percy Liang}, {and} \bibinfo{person}{Jure Leskovec}.}
  \bibinfo{year}{2022}\natexlab{}.
\newblock \bibinfo{title}{Deep Bidirectional Language-Knowledge Graph
  Pretraining}.
\newblock
\newblock
\showeprint[arxiv]{2210.09338}~[cs.CL]
\urldef\tempurl%
\url{https://arxiv.org/abs/2210.09338}
\showURL{%
\tempurl}


\bibitem[Ye et~al\mbox{.}(2024)]%
        {ye2024languagegraphneeds}
\bibfield{author}{\bibinfo{person}{Ruosong Ye}, \bibinfo{person}{Caiqi Zhang},
  \bibinfo{person}{Runhui Wang}, \bibinfo{person}{Shuyuan Xu}, {and}
  \bibinfo{person}{Yongfeng Zhang}.} \bibinfo{year}{2024}\natexlab{}.
\newblock \bibinfo{title}{Language is All a Graph Needs}.
\newblock
\newblock
\showeprint[arxiv]{2308.07134}~[cs.CL]
\urldef\tempurl%
\url{https://arxiv.org/abs/2308.07134}
\showURL{%
\tempurl}


\bibitem[Zhang et~al\mbox{.}(2024a)]%
        {zhang2024unveilingimpactcodingdata}
\bibfield{author}{\bibinfo{person}{Xinlu Zhang}, \bibinfo{person}{Zhiyu~Zoey
  Chen}, \bibinfo{person}{Xi Ye}, \bibinfo{person}{Xianjun Yang},
  \bibinfo{person}{Lichang Chen}, \bibinfo{person}{William~Yang Wang}, {and}
  \bibinfo{person}{Linda~Ruth Petzold}.} \bibinfo{year}{2024}\natexlab{a}.
\newblock \bibinfo{title}{Unveiling the Impact of Coding Data Instruction
  Fine-Tuning on Large Language Models Reasoning}.
\newblock
\newblock
\showeprint[arxiv]{2405.20535}~[cs.AI]
\urldef\tempurl%
\url{https://arxiv.org/abs/2405.20535}
\showURL{%
\tempurl}


\bibitem[Zhang et~al\mbox{.}(2024b)]%
        {zhang2024chainpreferenceoptimizationimproving}
\bibfield{author}{\bibinfo{person}{Xuan Zhang}, \bibinfo{person}{Chao Du},
  \bibinfo{person}{Tianyu Pang}, \bibinfo{person}{Qian Liu},
  \bibinfo{person}{Wei Gao}, {and} \bibinfo{person}{Min Lin}.}
  \bibinfo{year}{2024}\natexlab{b}.
\newblock \bibinfo{title}{Chain of Preference Optimization: Improving
  Chain-of-Thought Reasoning in LLMs}.
\newblock
\newblock
\showeprint[arxiv]{2406.09136}~[cs.CL]
\urldef\tempurl%
\url{https://arxiv.org/abs/2406.09136}
\showURL{%
\tempurl}


\bibitem[Zhu et~al\mbox{.}(2024)]%
        {zhu2024investigatinginstructiontuninglarge}
\bibfield{author}{\bibinfo{person}{Kerui Zhu}, \bibinfo{person}{Bo-Wei Huang},
  \bibinfo{person}{Bowen Jin}, \bibinfo{person}{Yizhu Jiao},
  \bibinfo{person}{Ming Zhong}, \bibinfo{person}{Kevin Chang},
  \bibinfo{person}{Shou-De Lin}, {and} \bibinfo{person}{Jiawei Han}.}
  \bibinfo{year}{2024}\natexlab{}.
\newblock \bibinfo{title}{Investigating Instruction Tuning Large Language
  Models on Graphs}.
\newblock
\newblock
\showeprint[arxiv]{2408.05457}~[cs.CL]
\urldef\tempurl%
\url{https://arxiv.org/abs/2408.05457}
\showURL{%
\tempurl}


\end{thebibliography}

\appendix

\section{One Shot Example for Different Prompt Formats} \label{prompt-format}
\subsection{Explanation for Natural Language Prompt}
\begin{lstlisting}
The composer of It Goes Like It Goes is David Shire. The spouse of David Shire is Didi Conn. The spouse of the composer of It Goes Like It Goes is _
\end{lstlisting}

\subsection{JSON structure Prompt}
\begin{lstlisting}
{
    "composer": {
        "It Goes Like It Goes": "David Shire" 
    },
    "spouse": {
        "David Shire": "Didi Conn" 
    }
}
\end{lstlisting}

\subsection{Python Code Snippet for Prompt V1}
\begin{lstlisting}
# Step 1. Define relationships with explicit types
relationships = {
    'composer': {
        'It Goes Like It Goes': 'David Shire'  # It Goes Like It Goes is related to David Shire via relationship composer
    },
    'spouse': {
        'David Shire': 'Didi Conn'  # David Shire is related to Didi Conn via relationship spouse
    }
}

# Define entities and relationships
e1 = 'It Goes Like It Goes'
r1 = 'composer'
r2 = 'spouse'

# Step 2. (r1, e1) -> e2
e2 = relationships[r1][e1]

# Step 3. (r2, e2) -> e3
e3 = relationships[r2][e2]

# Output the result
print(f"{r2} of {r1} of {e1} is {e3}")

# when you run the code, it will output:
# The spouse of the composer of It Goes Like It Goes is Didi Conn
\end{lstlisting}

\subsection{Python Code Snippet for Prompt V2}
\begin{lstlisting}
    # Step 1. Define relationships with knowledge base
    class KnowledgeBase:
        def __init__(self):
            # Initialize an empty dictionary to store facts.
            # Each key is a tuple (entity1, relation), and the value is the entity2 related to entity1 through relation.
            self.facts = {{}}

        def add_fact(self, entity1, relation, entity2):
            # Add a fact to the knowledge base.
            # :param entity1: The starting entity.
            # :param relation: The relation from entity1 to entity2.
            #:param entity2: The related entity reached via the relation.

            self.facts[(entity1, relation)] = entity2

        def infer(self, entity, *relations):
            #Infer the resulting entity by traversing the relations starting from the given entity.

            #:param entity: The starting entity.
            #:param relations: A chain of relations to traverse.
            #:return: The resulting entity after applying the relations, or None if no such path exists.

            current_entity = entity
            for relation in relations:
                key = (current_entity, relation)
                if key in self.facts:
                    current_entity = self.facts[key]
                else:
                    # If the path does not exist, return None.
                    return None
            return current_entity

    # Example usage:
    # Create a knowledge base instance.
    kb = KnowledgeBase()

    # Step 2. Define entities and relationships
    e1 = 'It Goes Like It Goes'
    r1 = 'composer'
    e2 = 'David Shire'
    r2 = 'spouse'
    e3 = 'Didi Conn'

    # Add entities and relationships to the knowledge base.
    kb.add_fact(e1, r1, e2)
    kb.add_fact(e2, r2, e3)

    # Step 3. Perform inference.
    result1 = kb.infer(e1, r1)          # Should return David Shire, (It Goes Like It Goes, composer) -> David Shire
    result2 = kb.infer(e2, r2)          # Should return Didi Conn, (David Shire, spouse) -> Didi Conn
    result3 = kb.infer(e1, r1, r2)      # Should return Didi Conn, (It Goes Like It Goes, composer) -> David Shire, (David Shire, spouse) -> Didi Conn
    
    # Output the result
    print(result1)  # Output: David Shire is related to It Goes Like It Goes through composer
    print(result2)  # Output: Didi Conn is related to David Shire through spouse
    print(result3)  # Output: Didi Conn is related to It Goes Like It Goes through composer and spouse
\end{lstlisting}

\end{document}